\documentclass[conference]{IEEEtran}
\IEEEoverridecommandlockouts
\usepackage{cite}
\usepackage{amsmath,amssymb,amsfonts}
\usepackage{algorithmic}
\usepackage{graphicx}
\usepackage{textcomp}
\usepackage{xcolor}
\usepackage{ dsfont }

\def\BibTeX{{\rm B\kern-.05em{\sc i\kern-.025em b}\kern-.08em
    T\kern-.1667em\lower.7ex\hbox{E}\kern-.125emX}}
\begin{document}


\title{Looking back at Labels: A Class based Domain Adaptation Technique\\
\thanks{\textbf{Project:} \textit{https://vinodkkurmi.github.io/DiscriminatorDomainAdaptation}}
}

\author{\IEEEauthorblockN{Vinod Kumar Kurmi}
\IEEEauthorblockA{{Indian Institute of Technology Kanpur}\\
\tt\small vinodkk@iitk.ac.in}
\and
\IEEEauthorblockN{Vinay P. Namboodiri}
\IEEEauthorblockA{{Indian Institute of Technology Kanpur}\\
\tt\small vinaypn@iitk.ac.in}
}
\maketitle
\begin{abstract}
\noindent In this paper, we solve the problem of adapting classifiers across domains. We consider the problem of domain adaptation for multi-class classification where we are provided a labeled set of examples in a source dataset and we are provided a target dataset with no supervision. In this setting, we propose an adversarial discriminator based approach. While the approach based on adversarial discriminator has been previously proposed; in this paper, we present an informed adversarial discriminator. Our observation relies on the analysis that shows that if the discriminator has access to all the information available including the class structure present in the source dataset, then it can guide the transformation of features of the target set of classes to a more structure adapted space. Using this formulation, we obtain state-of-the-art results for the standard evaluation on benchmark datasets. We further provide detailed analysis which shows that using all the labeled information results in an improved domain adaptation.
\end{abstract}


\section{Introduction}

Deep learning frameworks have solved many computer vision tasks such as object recognition, object detection, image generation, etc. With the advent of deep learning, models that are trained on a large number of images are ubiquitously being used. However, it was shown by Tzeng {\it et al.}~\cite{tzeng_arxiv2014} that while generically trained deep networks have a reduced dataset bias, there still exists a domain shift between different datasets and it is required to adapt the features appropriately. 
In the adversarial framework, one of the methods viz. unsupervised domain adaptation through backpropagation~\cite{ganin_ICML2015}, solved this problem by adding an auxiliary task that solves the problem of domain classification.  The main observation for this method is that for classifiers to be adapted across domains, the domain classifier should fail. This can be easily achieved through a gradient reversal layer that modifies features to worsen the ability to classify domains and it requires no labels to be available in the target dataset.
But, due to the limited capacity of using a binary discriminator, it introduces a problem of mode collapse in the feature space for source and target domains. The binary discriminator tends to mix all the target (or source) samples into a single domain class. In contrast to a binary discriminator, the proposed informative discriminator considers all the source label information and encourages target samples to be mis-classified into one of the source class.
It helps the target sample to preserve its multiple modes. For the source sample, the multiple modes are preserved by the regular classifier. The binary discriminator model\cite{ganin_ICML2015} is considered as a baseline for the proposed model. Through the proposed method we show that by making the adversarial domain classifier informative and providing it all the information available at the source, one can obtain an improved performance achieving an impressive improvement of 8.2\% over the corresponding baseline of gradient reversal~\cite{ganin_ICML2015} on the Amazon-DSLR adaptation task. This method also obtains an improvement of 6.21\% over the very recently proposed approach that also considers introducing additional source label information~\cite{pei_aaai2018multi}. 

\noindent To summarize, this paper makes the following contributions:
\begin{itemize}
    \item We propose a method that uses all source label information in a scalable way by using an informative domain discriminator.
    \item We show that providing the source label in the discriminator helps to preserve the mode information of target samples.
    \item This paper also provides additional insights into understanding our method by providing results for hyper-parameter sensitivity, inclusion of hierarchical class labels, discrepancy distance, statistical significance tests and feature visualization. This level of detailed analysis comprehensively supports our claims regarding the efficacy of the proposed approach.
\end{itemize}
\section{Literature Review}
\begin{figure*}[ht]
\centering
\includegraphics[height=7cm,width=14cm]{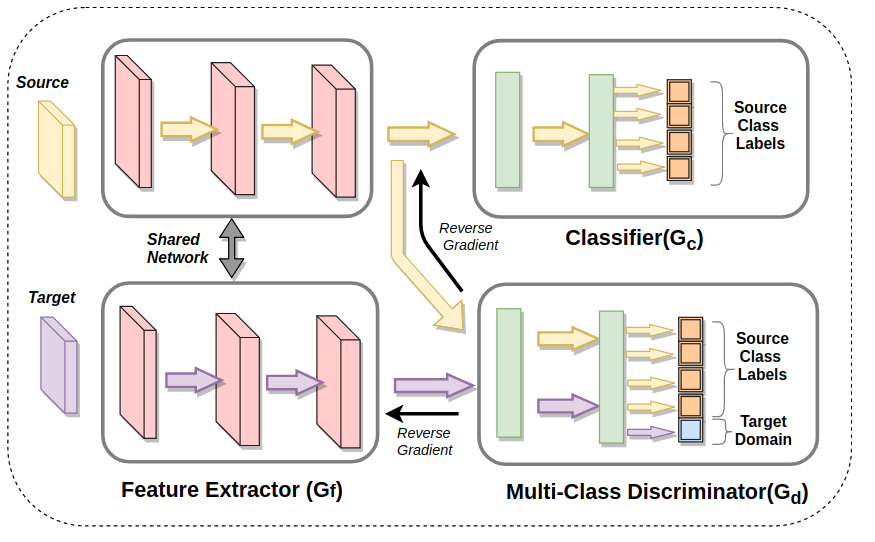}
\caption{The proposed architecture includes a deep feature extractor ($G_f$), classifier ($G_c$) and discriminator ($G_d$).}
\label{fig:univerise}
\end{figure*}
Domain adaptation problem has been widely studied in the research area of computer vision. 
Many successful domain adaptation methods try to make the source and target distribution closer to each other. Tzeng {\it et al.}~\cite{tzeng_arxiv2014} uses the maximum means discrepancy (MMD) in the feature space of source and target domain.  The multi kernel MMD between other layers applied in the Domain adaptation network (DAN)~\cite{long_ICML2015}. Other MMD based methods are proposed such as joint adaptation network (JAN)~\cite{long_icml2017deep} and~\cite{yan_CVPR2017}. Target entropy minimization used in the residual domain adaptation network (RTN)~\cite{long_NIPS2016}. Correlation alignment based domain adaptation achieved by the minimizing the Coral loss~\cite{sun_DACVA2017}. The coral loss based method also applied in the deep learning framework called Deep Coral~\cite{sun_ECCV2016}. Other coral loss based method proposed in the~\cite{sun_2016AAAI}. In the~\cite{shen_AAAI2018wasserstein} Wasserstein distance~\cite{arjovsky_icml2017wasserstein} has been applied to make the two distribution closer. Other similarity based adaptation network such as associative domain adaptation network~\cite{haeusser_iccv2017associative} also has been successfully applied. The subspace alignment based model was proposed by the~\cite{fernando_ICCV2013}. In the~\cite{pinheiro_cvpr2017unsupervised} model learns a pairwise similarity function between the classes to solve the domain adaptation problem. Image generation task  using the  Generative adversarial network \cite{goodfellow_NIPS2014}, applied in the unsupervised domain adaptation task. In the~\cite{ghifary_ICCV2015}, auto encoder framework used to learn the source to target mapping. Recent work by~\cite{sankaranarayanan_cvpr2018learning,murez_cvpr2018image,bousmalis_CVPR2017,choi2017_cvprstargan,Sankaranarayanan_CVPR2018,Liu_cvpr2018} used the similar concept by the generative adversarial network to adapt the target domain. For source to target mapping, cycle consistency framework~\cite{hoffman_iclr2018cycada}  for the domain transfer is proposed by the \cite{zhu_CVPR2017}. Multimodal text generation problem from the image and other modality, has been studied in~\cite{patro2018multimodal}. In the asymmetric domain adaptation models, different feature extractors for source and target have been used~\cite{bousmalis_nips2016domain,rozantsev_PAMI2018,saito_icml2017asymmetric,rozantsev_cvpr2018residual}. In the~\cite{li_iclr2016revisiting} different batch normalization used for the adaptation. Other work such as~\cite{carlucci_ICCV2017} used the different alignment layer between the source and target network. \cite{ saito_cvpr2017maximum} propose to maximize the discrepancy between two classifiers’s outputs to detect target samples that are far from the support of the source. Adversarial dropout~\cite{saito_arxiv2017}  was applied to adapt the target domain. Recently attention based model has been proposed in the~\cite{Kang_eccv2018} to solve the domain adaptation problem. Other exemplar based method~\cite{patro2018differential} have also explored ways to bring similar attention distributions closer.
There are other frameworks based on adversarial learning used to solve the domain adaptation problem. The proposed model lies in the adversarial domain adaptation framework setting. \cite{ganin_ICML2015} used a simple binary discriminator to learn indistinguishable feature mapping by the discriminator between the source and target domain. There are other adversarial method such as Adversarial discriminator for domain adaptation (ADDA)~\cite{tzeng_CVPR2017}, conditional adversarial domain adaptation (CDAN)~\cite{long_nips2018conditional}, Multi discriminator for domain adaptation (MADA)~\cite{pei_aaai2018multi}, Cycle-Consistent Adversarial Domain Adaptation (CyCDA)~\cite{hoffman_iclr2018cycada}, PixelDA \cite{bousmalis_CVPR2017}, PADA~\cite{Cao_eccv2018} and other recent adversarial feature learning methods are proposed in the~\cite{li_cvpr2018domain,saito_arxiv2017,zhang_cvpr2018collaborative,volpi_cvpr2017adversarial,chen_cvpr2018re,zhang_cvpr2018importance}
Domain adaptation in scene graph was also applied in~\cite{kumar2019adversarial}. Recent work by \cite{saito_cvpr2017maximum} adapts the classifier by maximizing the discrepancy between two classifier's outputs to detect target samples that are far from the support of the source. This method uses the prediction probability of the target samples to measure the discrepancy. The closest related work to our approach is the work by~\cite{pei_aaai2018multi} that extends the gradient reversal work~\cite{ganin_ICML2015} by including a class specific  discriminator. 
Class structure based adaptation has been explored by CDAN~\cite{long_nips2018conditional}. CDAN uses the class structure to compute the focal loss by concatenating the class prediction with features. But our objective is to make the discriminator more efficient by providing the sub class
structure~\cite{luo_IVS2008,hoffmann_ECML2001}. So the idea of using the class label structure is different than that of CDAN. Specifically, in the discriminator, we add more information regarding the source classes (the negative class of the discriminator). Other than this, CDAN and MADA both use the predicted target label (from the classifier) to compute the conditional distribution. But in our case, we are not relying on the target prediction score. Our method is complementary to most of the advances made in adversarial techniques and shows that an informative discriminator is crucial for obtaining significant improvements in the adversarial setting.

\section{Background: Discriminator for Domain Adaptation}
 The seminal generative adversarial network (GAN)~\cite{goodfellow_NIPS2014} and its different variants, used an adversarial loss to make a generator learn the true data distribution. The basic motivation of the adversarial methods~\cite{goodfellow_NIPS2014} is to align the fake (generated) and real distributions. These, however do so without considering the complex multimodal structures underlying in these data distributions. As a result, all the generated data classes are confused with real data. It leads to loss of the discriminative structure of data for different distribution. The discriminative structure  from the discriminator also helps to generate the paraphrase sentence generation problem~\cite{patro2018learning}. Particularly, in the domain adaptation scenario, it is crucial to preserve the multimodal structured data for solving the classification problem. To overcome the mode collapse problem, one of the solutions is proposed by Odena {\it et al.}\cite{odena_ICML2017} by extending the vanilla GAN architecture by introducing the auxiliary classification task on the discriminator. It leads to generated images being more distinct and sharp. Our proposed discriminator considers all the source label information to prevents loss of modes. We also propose a hierarchical structure for the discriminator when suitable. All the adversarial methods performance relies on the efficiency of the discriminator. One could make it more efficient by providing the sub-class structure of the data as suggested by~\cite{luo_IVS2008,hoffmann_ECML2001}. In the discriminator classification task, class labels can be considered as sub-classes, and they are well defined and distinct. By providing it to the discriminator, we observe that it helps to learn additional structure of the data.

 
 \subsection{ Domain-Label Discriminator }
Domain label discriminator is a simple binary classifier, which aims to misclassify the source and target domain samples. Most of the adversarial models~\cite{ganin_ICML2015, arjovsky_icml2017wasserstein} use the domain label discriminator for the image generation and domain adaptation. In the binary discriminator, all target samples (or generated fake samples) are mis-classified as a single source (true) domain. In the domain adaptation scenario, the target features generated by the feature extractor, obtains invariance in domain, but loses its multimodal structure. This mode information can be preserved by the proposed informative discriminator.
\subsection{ Class-Label Discriminator }
If two distributions are underlying the modes, then performance of a binary classifier can be improved by using the sub- class structure of dataset~\cite{luo_IVS2008,hoffmann_ECML2001}. In the proposed model, we use a class label based discriminator to improve the capacity of the discriminator to classify the source and target distribution.  
\cite{odena_ICML2017} applied an auxiliary classifier, in the discriminator of the  vanilla GAN architecture~\cite{radford_arxive2015unsupervised,goodfellow_NIPS2014}, to predicts the class label for generate better images without mixing the multimodal structure of data. These works provide an intuition that by providing the information about the structure of data to discriminator, data distribution can be captured effectively. To keep the multi-modal structure of target data MADA~\cite{pei_aaai2018multi} used $N$-discriminators for each mode (class) of data. The underlying problem with such multiple discriminator approaches is that its scalability is limited. The model will be more complex as the number of classes increases.  Another problem with MADA is that it uses the target class prediction score to decide the discriminator. For a wrong prediction it may lead to a different mode. We instead use a single discriminator to perform the source and target classification task along with the task of predicting the correct source label.


\section{Proposed Approach}
\subsection{Problem Description}
We address the problem of unsupervised domain adaptation, where there are no labels in the  training data for the target domain.
More formally, we are given data for a source
domain, $S = {(x_i^s,y_i^s)}_{i=1}^{n_s}$ of $n_s$ labeled examples and a target
domain, $T = {(x_i^t)}_{i=1}^{n_t}$  of $n_t$ unlabeled examples. The labels are not provided. 
The source domain and target domain are sampled from
joint distributions $P(X_s,Y_s)$ and $Q(X_t,Y_t)$ respectively, where $P \neq Q$. The assumption here is that the label set is common for source and target domains. 
The aim of our model is to provide a deep neural network that enables learning of transferable features $f = G_f(x)$ and an adaptive classifier $y = G_y(f)$ to reduce the
shift in the joint distributions across domains, such that the
target risk $Pr_{(x,y)\sim q }[G_y(G_f(x)\neq y]$ is minimized by jointly
minimizing the source risk and the distribution discrepancy by a discriminated
domain adaptation. At the time of training, we have access to all the source domain data  along with corresponding labels $S={(x_i^s,y_i^s)}_{i=1}^{n_s}$ and all unlabeled target data $T={(x_i^t)}_{i=1}^{n_t}$.

\subsection{Proposed Model}
In this proposed model, there are three main components: feature extractor ($G_f$), classifier ($G_c$) and informative discriminator ($G_d$). All the components are deep neural networks. This model is trained in an end-to-end fashion using the adversarial and classification loss. In contrast to~\cite{ganin_ICML2015} and~\cite{pei_aaai2018multi}, instead of a binary discriminator, we use a multi class discriminator. We also do not use any prediction score to predict or adapt the target samples, and we empirically show that it is actually detrimental in the proposed model.


\subsubsection{Feature Extractor ($G_f$)}
Feature extractor is a deep feed forward  convolution neural network architecture consisting of different feed forward layers. The task of this module is to map the input data $x$ in the feature space $G_f(x)$. It is parameterized by parameter $\theta_f$. We assume that input data $x$ is mapped to a $D$-dimensional feature vector
$G_f(x,\theta_f) \in \mathds{R}_D$
\subsubsection{Classifier Network ($G_c$)}
Classifier is also a deep feed forward network, consisting of fully connected layers. It is parameterised by the $\theta_c$.  In the training time, it maps the source data feature $f_s$ obtained from the feature extractor to class label $y$. 
$G_c(f_s,\theta_c) \sim {Y}$ where  $\mathds{Y} $ is the  source class label distribution.
This module is trained based only on the cross-entropy loss between the predicted source label and the ground truth source label.
\subsubsection{Discriminator Network ($G_d$)}
The task of any discriminator is to learn the source and target discrepancy. It considers the source features $f_s$ or target features $f_t$ and help to mis-classify them as source or target label. In the proposed model, we used a multi-class discriminator, which maps the source sample to its class label value while target samples are classified as fake label. This module is parameterized by $\theta_d$.
Using the reverse gradient layer, a target sample will be mis-classified as one of the source class labels. In the case of source sample, all the source samples may also be mis-classified. But the loss of multi-modal structure for source is prevented by the classifier.


\subsection{Training and Loss Function}
During training, there are two objectives for the model, first is to minimize the label prediction loss on the source dataset and optimize the parameters of feature extractor and classifier. Other is to make features more indistinguishable for source and target domain. For making the features indistinguishable, we reverse the gradient from the discriminator to back-propagate to the feature extractor. Our discriminator trains to classify the source class label and target domain. We reverse the gradient to make the features that are unfavorable for the discriminator. 

The additional loss will encourage a target sample to be mis-classified as one of the source classes by the discriminator. In the case of binary discriminator the target sample will be mis-classified as source domain (mixing of all the classes). In contrast to the binary discriminator methods, in the proposed method, the classes will not be mixed up and will be classified as only one of the source class. 
For all the source samples, the discriminator mis-classifies into a single target domain, but this is already prevented by the classification module that is based on true labels for the source samples.
So by the proposed model both target and source sample features are prevented from having a mode collapse.
\begin{equation}
\begin{split}
loss (\theta_f,\theta_y, \theta_d) = & \frac{1}{n_s}\sum_{x_i \in D_s} L_y(G_y(G_f(x_i)),y_i) + 
                                \\ & \frac{\lambda}{n_s+n_t}\sum_{x_i \in D_s \cup D_t} L_d(G_d(G_f(x_i)),d_i)
\end{split}
\end{equation} 
where \begin{equation}
  d_i=\begin{cases}
    y_i, & \text{if $x_i \in D_s $}.\\
    |C|+1, & \text{ if $x_i \in D_t$ }.
  \end{cases}
\end{equation}

\noindent $\lambda$ is a trade-off parameter between the two objectives and $|C|$ is the number  of source classes. 
$L_y$ and $L_d$ are the cross entropy loss for classifier and discriminator respectively. $ D_s$ and $ D_t $ are the source and target domain respectively. $G_f,G_c$ and $G_d$ are the feature extractor, classifier and discriminator.

\section{Results \& Experiments}
In this section, we evaluate our model on various widely used benchmarks. Following the common setting in unsupervised domain adaptation, we used the Alexnet~\cite{krizhevsky_NIPS2012} architecture pre-trained on the Imagenet dataset for our base model. Results are compared with the state-of-art methods such as~\cite{tzeng_arxiv2014,ganin_ICML2015,long_ICML2015,sun_ECCV2016,ghifary_ECCV2016,long_NIPS2016,long_icml2017deep,yan_CVPR2017,carlucci_ICCV2017,murez_cvpr2018image,long_nips2018conditional, venkateswara_cvpr2017deep, pei_aaai2018multi}. The proposed model has been evaluated on the Office-31~\cite{saenko_ECCV2010}, Office-Home datasets~\cite{venkateswara_cvpr2017deep}, Caltech-Bing datasets~\cite{bergamo_NIPS2010} and ImageCLEF datasets. Other details and codes are provided in the project page \textit{https://vinodkkurmi.github.io/DiscriminatorDomainAdaptation}.


\begin{table*}
\centering
\caption{Classification accuracy evaluation of different domain adaptation approaches on the
standard OFFICE-31~\cite{saenko_ECCV2010} dataset. All methods are evaluated in the "fully-transductive" protocol using the Alexnet pretrained~\cite{krizhevsky_NIPS2012} model. Our method (last row) outperforms competitors on the three adaptation task.} \label{office_table}
\scalebox{1.2}{
\begin{tabular}{ |p{1.8cm}|c|c|c|c|c|c|c| } 
 \hline
 Method & A $\rightarrow$ W & D $\rightarrow$ W &  W $\rightarrow$ D  &  A $\rightarrow$ D  &  D $\rightarrow$ A  &  W $\rightarrow$ A & Avg\\ 
  \hline
DDC\cite{tzeng_arxiv2014} & 61.0 $\pm$ 0.5 & 95.0 $\pm$ 0.3  & 98.5 $\pm$ 0.3   & 64.9 $\pm$ 0.4 & 47.2 $\pm$ 0.5 & 49.4 $\pm$ 0.4  & 69.3 \\ 
DAN\cite{long_ICML2015} & 68.5 $\pm$ 0.3 & 96.0 $\pm$ 0.1 & 99.0 $\pm$ 0.1  & 66.8 $\pm$ 0.2   & 50.0 $\pm$ 0.4 & 49.8 $\pm$ 0.3  &71.6 \\ 
 DeepCoral\cite{sun_ECCV2016} & 66.4 $\pm$ 0.4 & 95.7 $\pm$ 0.3 & 99.2 $\pm$ 0.1  & 66.8 $\pm$ 0.6   & 52.8 $\pm$ 0.2 & 51.5 $\pm$ 0.3  & 72.0 \\ 
  WDAN\cite{yan_CVPR2017} & 66.9 $\pm$ 0.2 & 95.9 $\pm$ 0.2 & 99.0 $\pm$ 0.1  &64.4 $\pm$ 0.2  & 53.8 $\pm$ 0.1 & 52.7 $\pm$ 0.2 & 72.1 \\ 

  DHN\cite{venkateswara_cvpr2017deep} & 68.3 $\pm$ 0.0 & 96.1 $\pm$ 0.0 & 98.8 $\pm$ 0.0  & 66.4 $\pm$ 0.0 &  55.5 $\pm$ 0.0 &  53.0 $\pm$ 0.0 & 73.0 \\ 
DRCN\cite{ghifary_ECCV2016} & 68.7 $\pm$ 0.3 & 96.4 $\pm$ 0.3  & 99.0 $\pm$ 0.2  & 66.8 $\pm$ 0.5  & 56.0 $\pm$ 0.5 & 54.9 $\pm$ 05 & 73.6\\ 
 RTN\cite{long_NIPS2016} & 73.3 $\pm$ 0.2 & 96.8 $\pm$ 0.2 & 99.6 $\pm$ 0.1 & 71.0 $\pm$ 0.2 & 50.5 $\pm$ 0.3 & 51.0 $\pm$ 0.1  & 73.7 \\ 
GRL\cite{ganin_ICML2015} & 73.0 $\pm$ 0.5 & 96.4 $\pm$ 0.3  & 99.2 $\pm$ 0.3  & 72.3 $\pm$ 0.3  & 52.4 $\pm$ 0.4 & 50.4 $\pm$ 0.5  & 73.9 \\ 
 I2I\cite{murez_cvpr2018image} & 75.3 $\pm$ 0.0 & 96.5 $\pm$ 0.0 &  99.6 $\pm$ 0.0  &71.1 $\pm$ 0.0 & 50.1 $\pm$ 0.0 & 52.1 $\pm$ 0.0 & 74.1 \\ 
 JAN\cite{long_icml2017deep} & 75.2 $\pm$ 0.4 & 96.6 $\pm$ 0.2 & 99.6 $\pm$ 0.1  & 72.8 $\pm$ 0.3  & 57.5 $\pm$ 0.2 & 56.3 $\pm$ 0.2  & 76.3\\ 
 CDAN\cite{long_nips2018conditional} & 77.9 $\pm$ 0.3 & 96.9 $\pm$ 0.2 & 100.0 $\pm$ 0.0  & 74.6 $\pm$ 0.2 &  55.1 $\pm$ 0.3 &  57.5 $\pm$ 0.4 & 77.0 \\ 
  ADIAL\cite{carlucci_ICCV2017} & 75.5 $\pm$ 0.0 & 96.6 $\pm$ 0.0 &  99.5 $\pm$ 0.0  & 73.6 $\pm$ 0.0 &  \textbf{58.1 $\pm$ 0.0} &  \textbf{59.4 $\pm$ 0.0} & 77.1\\ 
 MADA\cite{pei_aaai2018multi} & 78.5 $\pm$ 0.2 & {99.8 $\pm$ 0.1} & 100.0 $\pm$ 0.0  & 74.1 $\pm$ 0.1 &  56.0 $\pm$ 0.2 &  54.5 $\pm$ 0.3 & 77.1 \\ 
 \hline
 IDDA[ours]& \textbf{82.2 $\pm$ 0.8} & \textbf{99.8 $\pm$ 0.2} &  \textbf{100.0 $\pm$ 0.0} &  \textbf{82.4 $\pm$ 0.5} & 54.1 $\pm$ 0.4  & 52.5 $\pm$ 0.3  &  \textbf{78.5}\\ 
 \hline
\end{tabular}
}
\end{table*}
\subsection{Office-31 dataset }
Office-31~\cite{saenko_ECCV2010} is a benchmark for domain adaptation, comprising 4,110 images in 31 classes collected from three distinct domains: Amazon (A), which contains images downloaded from amazon.com, Webcam (W) and DSLR (D), which contain images taken by web camera and digital SLR camera with different photographical settings, respectively. 
To enable unbiased evaluation, we evaluate on all the 6 possible transfer tasks such as A$\rightarrow$W, D$\rightarrow$ W, W$\rightarrow$D, A$\rightarrow$D, D$\rightarrow$A and W$\rightarrow$A.
The performance is shown in Table~\ref{office_table}.
It is noteworthy that the proposed model promotes the classification accuracy substantially on hard transfer tasks, e.g., A$\rightarrow $W and A$\rightarrow$D,  where the source and target domains are substantially different. In the 4 out of 6 shifts, it achieves the highest accuracy. For the other 2 shifts, we achieve comparable performance because in these cases the source domain has fewer examples than target domain. Therefore it becomes hard for the discriminator to learn the modes of the datasets for this case. However, we can see that average performance is better than all the other base line methods.  The encouraging results highlights the importance of informative discriminator based domain adaptation in deep neural networks, and suggests that this model is able to learn more transferable representations for effective domain adaptation.
\subsection{Home-Office dataset}
We also evaluated our model on the Office-Home dataset~\cite{venkateswara_cvpr2017deep} for unsupervised domain adaptation. This dataset consists of four domains, Art (Ar), Clipart (Cl), Product (Pr) and Real-World (Rw). Each domain has common 65 categories. 
The Art domain contains the artistic description of objects such as painting, sketches etc. The Clipart are the collection of clipart images. In the Product domain images have no background. The Real-World domain consists of object capture from the regular camera.
We evaluated our model by considering the Art data as  source data and remaining dataset as target dataset. So we have three adaptation tasks, Ar $\rightarrow$ Cl, Ar $\rightarrow$ Pr and Ar $\rightarrow$ Rw. The performance reported in Table~\ref{home_office}.


\begin{table}[]
\caption {Classification accuracy (\%) on Home-Office dataset~\cite{venkateswara_cvpr2017deep} for unsupervised domain adaptation on AlexNet~\cite{krizhevsky_NIPS2012} model} 
\begin{center}
  \centering
\begin{tabular}{ |c|c|c|c|c| }
 \hline
  \textbf{Method } & Ar $\rightarrow$ Cl & Ar $\rightarrow$ Pr &  Ar $\rightarrow$ Rw  & Avg \\ 
  \hline
 Alexnet\cite{krizhevsky_NIPS2012}  &  26.4  & 32.6  & 41.3   & 33.43 \\
     DAH\cite{venkateswara_cvpr2017deep}  &  31.6  & 40.7   & 51.7    & 41.33 \\
  DAN\cite{long_ICML2015}  & 31.7 & 43.2 & 55.1&    43.33  \\ 
   GKT\cite{ding_eccv2018graph}  & 34.5 & 43.6  &  55.3  & 44.46\\
  GRL\cite{ganin_ICML2015} & 36.4 & 45.2  & 54.7    &  45.43   \\
  JAN\cite{long_icml2017deep}  &  35.5  & 46.1   & 57.7    &  46.43  \\
 CDAN \cite{long_nips2018conditional} & 38.1 & 48.7  & \textbf{60.3}    & 49.03\\

 \hline
IDDA[ours]& \textbf{38.9} & \textbf{50.7} & 58.8  &  \textbf{49.46}    \\
 \hline
\end{tabular}
\end{center}
\vspace{-8mm}
\label{home_office}
 \end{table}

\begin{table}
\caption{Classification accuracy evaluation of different discriminator based domain adaptation approaches on the Caltech-Bing and MNIST \& MNIST-M dataset.}\label{cb_table}
\begin{center}
\begin{tabular}{ |p{3cm}|c|c|c| } 
 \hline
 Method & C $\rightarrow$ B & B $\rightarrow$ C & M $\rightarrow$ M-M  \\ 
  \hline

Source Only\cite{krizhevsky_NIPS2012} & 36.16  & 72.67  & 52.25 \\
Binary  Discriminator\cite{ganin_ICML2015} & 36.35 & 73.29 & 76.66 \\
 \hline
Parent Label Discriminator[our] & 36.50  & 73.87 &  - \\ 
 \hline
 Class Label Discriminator(IDDA)[ours]& \textbf{36.98} & \textbf{74.62} &\textbf{82.29}  \\ 
 \hline
\end{tabular}
\end{center}

\end{table}

\subsection{Caltech-Bing dataset}
For demonstrating the idea that if we provide more source dataset information to discriminator, it performs well, we used subset of Caltech-Bing dataset~\cite{bergamo_NIPS2010} that consists of 43 classes with 3 parents classes as aquatic (11 classes), terrestrial (23 classes) and  avian (9 classes). We call it mini-Bing (B) and mini-Caltech (C) dataset. There are total 4960 images in mini-Caltech and 20731 images in mini-Bing dataset. The performance on both task (Caltech $\rightarrow$ Bing and Bing $\rightarrow$ Caltech) are shown in Table~\ref{cb_table}.
 \subsection{MNIST-MNIST-M dataset }
 We also experimented with the MNIST dataset as source data. In order to obtain the target domain (MNIST-M)~\cite{lecun_ProceedingsIEEE1998} we blend digits from the original set over patches randomly extracted from color photos from BSDS500~\cite{arbelaez_PAMI2011}. The adaptation result is shown in the Table~\ref{cb_table}.
\subsection{Results on ImageCLEF Dataset}
ImageCLEF-2014 dataset consists of 3 domains: Caltech-256 (C), ILSVRC 2012 (I), and Pascal-VOC 2012 (P). There are 12 common classes, and each class has 50 samples.There is a total of 600 images in each domain.
We evaluate models on all 6 transfer tasks: I$\rightarrow$P, P$\rightarrow$I, I$\rightarrow$C, C$\rightarrow$I, C$\rightarrow$P, and P$\rightarrow$C. The results on the ImageCLEF are reported in Table~\ref{tbl:imageclef}. 
\vspace{-1em}
\begin{table}[!]
\centering
\caption {Classification accuracy (\%) on \textit{ImageCLEF} dataset for unsupervised domain adaptation (AlexNet~\cite{krizhevsky_NIPS2012})} 
\begin{tabular}{|p{1.3cm}|p{0.64cm}|p{0.64cm}|p{0.64cm}|p{0.64cm}|p{0.64cm}|p{0.65cm}|p{0.5cm}|}
 \hline
  \textbf{Method }& I$\rightarrow$P & P$\rightarrow$I &  I$\rightarrow$C &C$\rightarrow$I & C$\rightarrow$P & P$\rightarrow$C & Avg \\ 
 \hline
AlexNet~\cite{krizhevsky_NIPS2012} & 66.2 & 70.0 & 84.3 & 71.3 & 59.3 &84.5 & 73.9  \\
DAN\cite{long_ICML2015} & 67.3 & 80.5 & 87.7 & 76.0 & 61.6 & 88.4 & 76.9  \\
GRL~\cite{ganin_ICML2015} & 66.5 & 81.8 & 89.0 & 79.8 &63.5 &88.7 &78.2  \\
RTN\cite{long_NIPS2016} & 67.4 & 82.3 &89.5 & 78.0 &63.0 &90.1 &78.4  \\
MADA~\cite{pei_aaai2018multi} & 68.3 &\textbf{83.0} &91.0 &80.7 &63.8 & 92.2 &79.8  \\
 \hline
 IDDA & \textbf{68.3 }& 81.8 & \textbf{92.3} & \textbf{81.6} & \textbf{67.2} & \textbf{92.8} & \textbf{80.6}\\ 
 \hline
\end{tabular}
  \label{tbl:imageclef}
 \end{table}





\begin{figure*}[!htb]
\minipage{0.3\textwidth}
  \includegraphics[width=\linewidth]{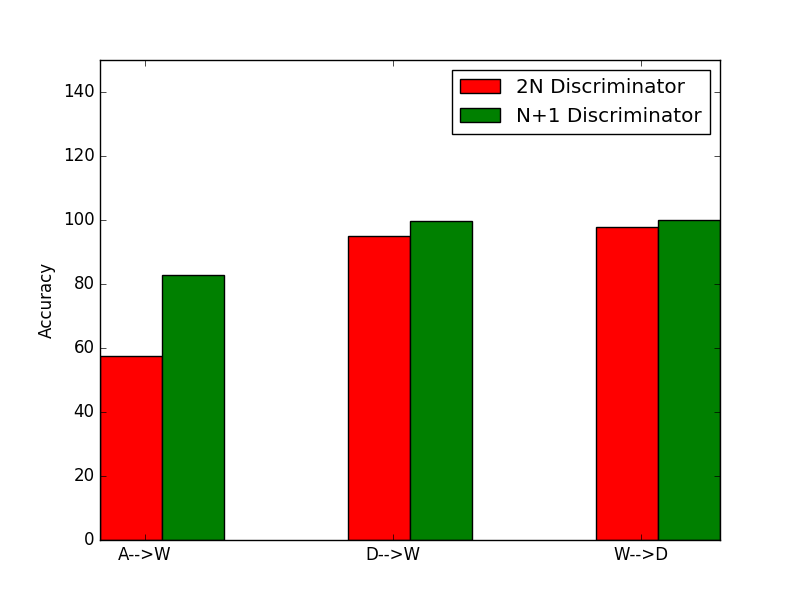}
 \caption{Classification accuracy on office dataset (A$\rightarrow$W, D$\rightarrow$W and W$\rightarrow$D), when we use the predicted target label(from the softmax of classifier) in the discriminator.} \label{2n_table}
\endminipage\hfill
\minipage{0.3\textwidth}
  \includegraphics[width=\linewidth]{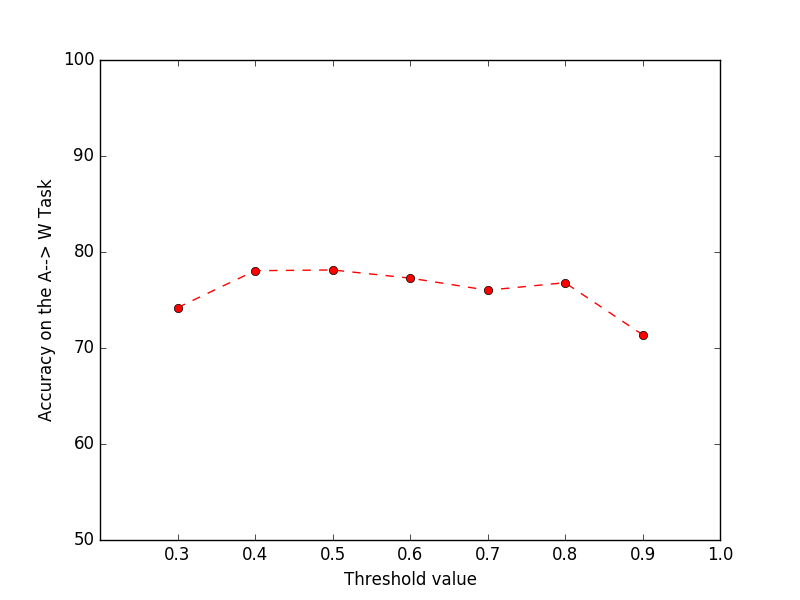}
   \caption{Classification accuracy on office (A$\rightarrow$W) dataset, when we consider the most confident predicted target label (from the softmax of classifier) to train the model.}\label{thre_table}
\endminipage\hfill
\minipage{0.3\textwidth}%
  \includegraphics[width=\linewidth]{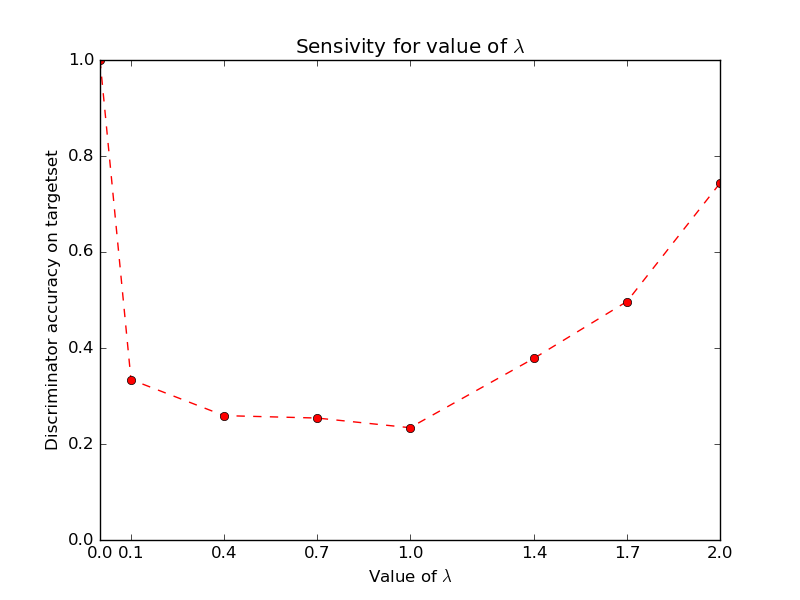}
  \caption{Sensitivity on the $\lambda$ experiment. Discriminator accuracy with respect to $\lambda$ on office (A$\rightarrow$W) dataset. Here  $\lambda=1.0$ gives the lowest accuracy , which is desired in the domain adaptation task} \label{fig:lamda}
\endminipage
\end{figure*}
\section{Analysis}

\begin{figure}
     \small
     \centering
     \begin{tabular}[b]{ c  c}
     \includegraphics[width=0.23\textwidth]{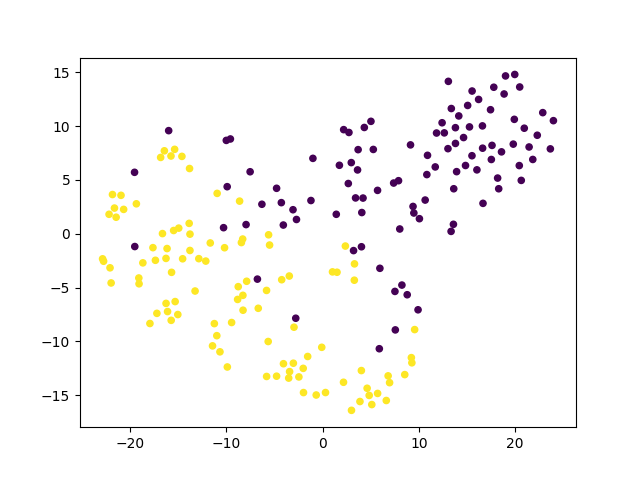}
     & \includegraphics[width=0.23\textwidth]{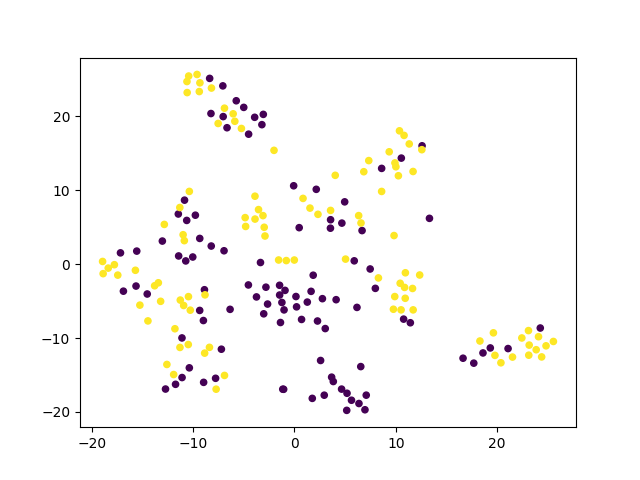}\\
          (a) Before Adaptation & (b) After Adaptation  \\ 
       \end{tabular}
      \caption{The effect of adaptation on the distribution of the extracted features (best viewed in color) for MNIST $\rightarrow$ MNIST-M dataset. The figure shows t-SNE visualizations of the CNN's activation (a) in case when no adaptation was performed and (b) in case when our adaptation procedure was incorporated into training. yellow points correspond to the source domain examples (MNIST data), while violet ones correspond to the target domain (MNIST-M). Adaptation makes the two distributions of features much closer}
      \label{tbl:tSNE}
      \vspace{-5mm}
 \end{figure}
 
 In this section we provide an analysis for the proposed  model using the aspects such as distribution discrepancy and statistical significant test. We further provide hyper-parameter sensitivity and feature visualization analysis.
 
 \vspace{-0.2cm}
\subsection{ Target risk error minimization}
\subsubsection{Domain adaptation theory}
As suggest by the~\cite{ben_ML2010} the target risk error  $\epsilon_t(h)$ for hypothesis $h$ is bounded by the source risk error and distribution distance.
\begin{equation} 
 \epsilon_t(h_1,h_2)  \leq  \epsilon_s(h_1,h_2) + \frac{1}{2} d_{\mathcal{H}{\Delta}\mathcal{H} }(\mathcal{D}_s, \mathcal{D}_t )
\end{equation}
 where the source-target distance is defiend as:
\begin{equation}
\begin{split}
&d_{\mathcal{H}{\Delta}\mathcal{H} (\mathcal{D}_s, \mathcal{D}_t )}\\
     &= 2 \sup_{h1,h2 \in H}| P_{\mathcal{D}_s} [ \ h(x)=1 ]|
     -| P_{\mathcal{D}_t} [ \  h(x)=1 ]|
     \end{split}
\end{equation}
where  $h(x)= h_1(x) \oplus h_2(x)$

\subsubsection{Symmetric hypothesis space for multi-class classification}
We can construct a symmetric difference hypothesis $\mathcal{H}_c{\Delta}\mathcal{H}_c $ \cite{blitzer_NIPS2008} for multi-class hypothesis. Choose hypothesis  $h(x)= h_1(x) \oplus h_2(x) \dots \oplus h_c(x) $. 
 $\mathcal{H}_c{\Delta}\mathcal{H}_c=\{h | h=h_1(x) \oplus h_2(x)\dots \oplus h_c(x)$ , $h_1,h_2 \dots h_c(x) \in \mathcal{H} \}$ where $c$ is the number of classes.
\noindent We assume that the hypothesis space $\mathcal{H}_c$ is the set of all hypothesis produced by the classifier. Similarly hypothesis space $\mathcal{H}_d$ is the set of all hypothesis produced by the discriminator. Consider fixed $\mathcal{D}_s$  and $\mathcal{D}_t$  over the representation space produced by the feature extractor  and a family of label predictors $\mathcal{H}_c$. 

\vspace{-0.3cm}
\begin{equation}
\begin{split}
d_{\mathcal{H}_c{\Delta}\mathcal{H}_c{\Delta}} (\mathcal{D}_s, \mathcal{D}_t )\\
= 2 \sup_{h \in \mathcal{H}_c{\Delta}\mathcal{H}_c}&| P_{\mathcal{D}_s} [ \ h(x)=1 ]| -| P_{\mathcal{D}_t} [ \  h(x)=1 ]| \\ 
     \end{split}
\end{equation}
Assume that the family of domain classifiers  $\mathcal{H}_d$ is rich enough to contain the symmetric difference hypothesis set of $\mathcal{H}_c$. 
In the proposed class label based discriminator this assumption not only holds but also achieves a more tighter bound than the binary discriminator used in\cite{ganin_ICML2015} 

\begin{equation*}
\begin{split}
    & d_{\mathcal{H}_c{\Delta}\mathcal{H}_c{\Delta}} (\mathcal{D}_s, \mathcal{D}_t ) \\ 
    & \leq 2 \sup_{h \in \mathcal{H}_d{\Delta}\mathcal{H}_d}  |P_{\mathcal{D}_s} [ \ h(x)=1 ]-  P_{\mathcal{D}_t} [ \  h(x)=1 ]| 
     \end{split}
\end{equation*}
 
\begin{equation}
     d_{\mathcal{H}_c{\Delta}\mathcal{H}_c{\Delta}} (\mathcal{D}_s, \mathcal{D}_t ) \leq 2 \sup_{h \in \mathcal{H}_d{\Delta}\mathcal{H}_d}|\alpha(h)-1 | 
\end{equation}

\noindent where $\alpha(h) = P_{\mathcal{D}_s} [ \ h(x)=1 ]  + P_{\mathcal{D}_t} [ \  h(x)=0 ]  $.

Now we construct symmetric hypothesis set $\mathcal{H}_d{\Delta}\mathcal{H}_d $ such that
 $ h'(x)= h(x)\oplus h_{c+1}(x)$
 and $ P_{\mathcal{D}_s} [ \ h_{c+1}(x)=1 ] = 0  $ $ \forall x  \in \mathcal{D}_s $
 \begin{equation*}
 \begin{split}
    \mathcal{H}_d{\Delta}\mathcal{H}_d = \{h | & h=h_1(x) \oplus h_2(x) \dots \oplus h_c(x) \oplus h_{c+1}, \\
    & h_1,h_2 \dots h_c(x),h_{c+1} \in \mathcal{H} \} 
      \end{split}
 \end{equation*}
 \begin{equation}
    \alpha(h) \leq  P_{\mathcal{D}_s \cup \mathcal{D}_t } [ \ h'(x)=1 ]  =   \alpha(h')
\end{equation}
\noindent where $c$ is the number of classes. The hypothesis $ h'$ is achieved by the class based domain discriminator model $G_d$. Thus, optimal discriminator gives the upper bound for $\mathcal{H}_c{\Delta}\mathcal{H}_c$ At the same time, back propagation of the reversed gradient changes the representation space so that $\alpha(G_d) $ becomes smaller effectively reducing $  d_{\mathcal{H}_c{\Delta}\mathcal{H}_c} (\mathcal{D}_s, \mathcal{D}_t ) $ and leading to the better approximation of $\epsilon_s(h)$  by $\epsilon_t(h)$.

\subsubsection{Distribution discrepancy}
\begin{figure}
     \small
     \centering
     \begin{tabular}[b]{ c  c }
     (a) A  $\rightarrow$ W & (b)A  $\rightarrow$ D  \\ 
     \includegraphics[width=0.22\textwidth]{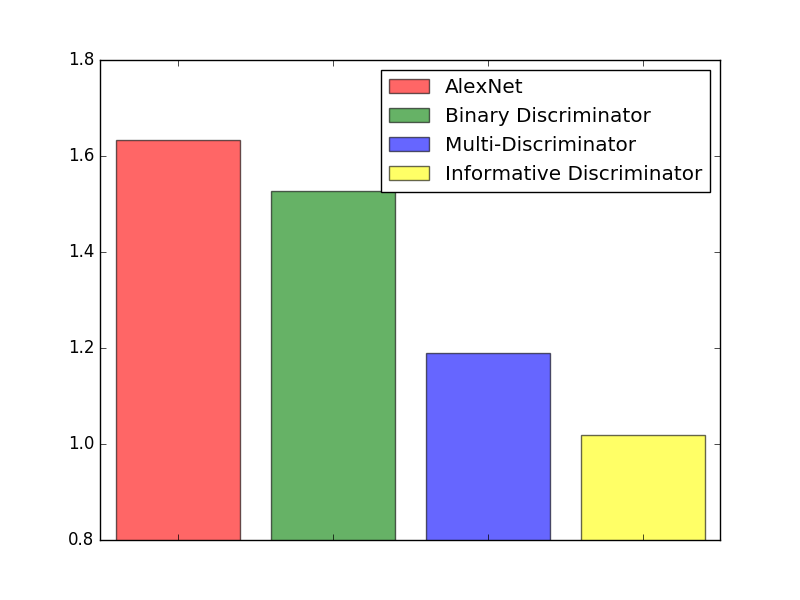}
     & \includegraphics[width=0.22\textwidth]{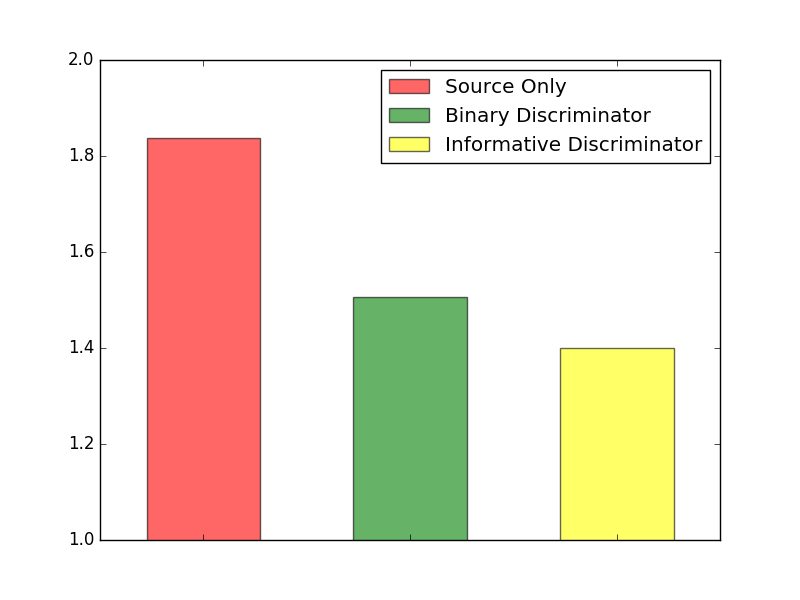}
       \end{tabular}
    \caption{Proxy $A$-distance for Amazon $\rightarrow$ Webcam and Amazon $\rightarrow$ DSLR tasks for method Source only, Binary discriminator~\cite{ganin_ICML2015} and proposed model. }\label{fig:proxy}
 \end{figure}

The domain adaptation theory~\cite{ben_ML2010} suggests $\mathcal{A}$-distance as a measure of cross domain discrepancy, which, together with the source risk, will bound the target risk. The proxy $\mathcal{A}$-distance is defined as $d_\mathcal{A} = 2 (1 - 2\epsilon)$, where $ \epsilon $ is the generalization error of a classifier (e.g. kernel SVM) trained on the binary task of discriminating source and target. Figure~\ref{fig:proxy} shows $d_\mathcal{A}$ on tasks A $\rightarrow $D and  A $\rightarrow $W, with features of source only model, Binary discriminator~\cite{ganin_ICML2015}, and proposed informative discriminator model. We observe that $d_\mathcal{A}$ using our model features is much smaller than  $d_\mathcal{A}$ using source only model and RevGrad(binary discriminator)~\cite{ganin_ICML2015} features, which suggests that our features can reduce the cross-domain gap more effectively.

\begin{figure*}[t]
     \small
     \begin{tabular}[b]{ c  c c}
     (a) A $\rightarrow$ D & (b) A $\rightarrow$ W  & (c) B $\rightarrow$ C  \\ 
     \includegraphics[width=0.27\textwidth]{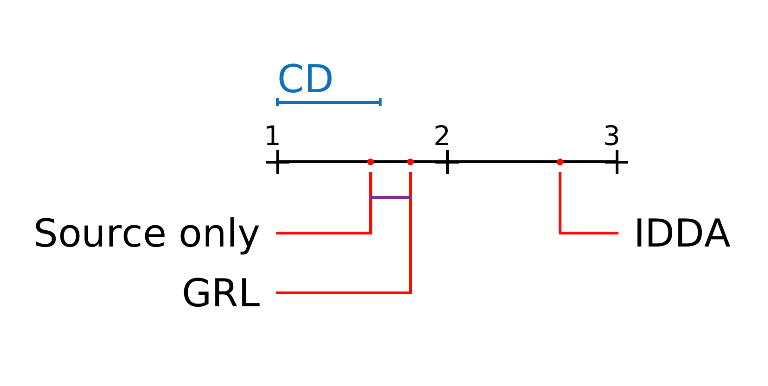}
     & \includegraphics[width=0.27\textwidth]{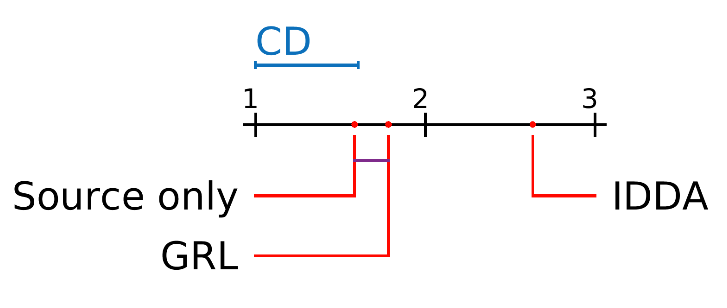}
      & \includegraphics[width=0.27\textwidth]{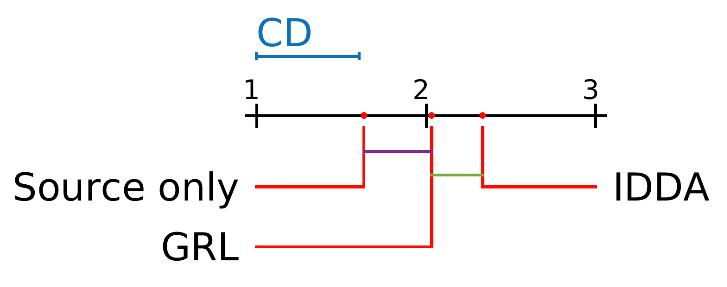}
       \end{tabular}
     \caption{Analysis of statistically significant difference for (a)A $\rightarrow$ D, (b)A $\rightarrow$ W and (c)B $\rightarrow$ C in Source only, Binary label Discriminator (GRL)~\cite{ganin_ICML2015} and proposed model (IDDA), with a significance level of 0.05. The mean rank is plotted on x-axis. The CD is 0.6051 and all the methods are way outside the CD, so are statistically significant over source only trained model. We can see IDDA is statistically significantly over GRL in all 3 adaptation tasks}\label{fig:ssa}.
 \end{figure*}
 
\subsection{Statistical significance analysis}
We have analysed statistical significance~\cite{demvsar_JMLR2006} for our proposed informative discriminator method against binary label discriminator~\cite{ganin_ICML2015} and source only method for the domain adaptation task. The Critical Difference (CD) for Nemenyi  test depends upon the given confidence level (which is 0.05 in our case) for average ranks and  number of tested datasets. If the difference in the rank of the two methods lies within CD (in our case CD = 0.6051), then they are not significantly different. Figure~\ref{fig:ssa} visualizes the post hoc analysis using the CD diagram for A $\rightarrow$ D,  A $\rightarrow$ W  and  B $\rightarrow$ C dataset respectively. From the figures, it is clear that our Informative discriminator (IDDA) is best and is significantly different from the GRL(binary discriminator)~\cite{ganin_ICML2015} and source only model.
\subsection{Feature visualization}
Adaptability  of target to source features can be visualized using the t-SNE embeddings of images feature. We follow similar setting in~\cite{tzeng_arxiv2014},~\cite{ganin_ICML2015} and~\cite{pei_aaai2018multi} and plot t-SNE embeddings of the MNIST $\rightarrow$ MNIST-M dataset in the Figures~\ref{tbl:tSNE}.
\subsection{Parameter sensitivity for discriminator accuracy}
We  also investigate the effects of the  value of parameter $\lambda  \in   \{0,0.1, 0.4, 0.7, 1, 1.4, 1.7, 2\}$.
We plot in Figure~\ref{fig:lamda} the discriminator accuracy on target data (classifying source v/s target domain) with respect to value of $\lambda $ on tasks Amazon $\rightarrow $ Webcam. Observe that discriminator accuracy first decreases and then increases as $\lambda$ varies and demonstrates a $U$-shaped curve. We choose $\lambda=1$ where the discriminator accuracy are lowest, i.e  source and target domain are more indistinguishable. In this experiment we do not use any target labels.

\subsection{Empirical evaluation of model by using predicted labels}
In a recent work~\cite{pei_aaai2018multi}, the authors considered the use of predicted target information. In order to validate the idea of using target class prediction label in the proposed model, we consider them in following ways:
\subsubsection{Using 2N class discriminator} 
 In this case we construct the discriminator for classifying is 2$N$ classes, where $N$ is the number of class label in the dataset. Here each sample is classify by the discriminator from the 2$N$ classes (it could belong to source class label or target class label). For the source data, we used the provided source data label, while in the case of target we used the soft-max output of target prediction probability from the classifier (C). Results for the three task of office dataset are shown in the Figure~\ref{2n_table}. We observed that use of predicted target information actually results in reduced performance.

 

\subsubsection{Using only confident target samples} 
We experimented for A $\rightarrow $W  by taking the target samples which are more certain about the class (using the classification softmax probability). We observed that in our cases avoiding use of target labels is better than using predictions. Figure~\ref{thre_table} shows the result for different threshold value for which, we consider that target sample to train the discriminator.
\vspace{-7.9mm}
\section{Conclusion}
We proposed a method for obtaining an informative discriminator that aids improved domain adaptation. Our analysis showed that this discriminator indeed helps us in obtaining statistically significant improvement that can also be justified theoretically. We further observed through visualization that domain adapted features do result in domain invariant feature representations. In future, we aim to further explore relations with respect to structured source representations that can yield improved domain adaptation. To some extent, we have already justified this through the use of hierarchical classifiers. The incorporation of structure in source and correlating that with the target structure is a promising direction which we have initiated through this work. 
\section{Acknowledgement}
The author Vinod Kumar Kurmi acknowledges support from TCS Research Scholarship Program.
{\small
\bibliographystyle{IEEEtran}
\bibliography{egbib.bib}
}

\end{document}